\pdfoutput=1
\PassOptionsToPackage{table}{xcolor}
\documentclass[11pt]{article}

\usepackage{acl}

\usepackage{rotating}
\usepackage{times}
\usepackage[table]{xcolor}
\usepackage{latexsym}
\usepackage{enumitem}
\usepackage{booktabs}
\usepackage{graphicx}
\usepackage{tablefootnote}
\usepackage{float}
\usepackage{array}
\usepackage{tikz}
\usetikzlibrary{shapes.geometric, arrows,backgrounds}
\usepackage{adjustbox}
\usepackage{multirow}
\usepackage{multicol}
\usepackage{amsmath} % for \sum command
\usepackage{amssymb} % for \mathbb{I} command

\tikzstyle{process} = [rectangle, minimum width=5.5cm,  minimum height=1cm, align=left, text width=5.4cm, draw=black, fill=orange!10]

\tikzstyle{arrow} = [thick,->,>=stealth]

\usepackage[T1]{fontenc}
\usepackage[utf8]{inputenc}

\usepackage{microtype}

\usepackage{inconsolata}

\title{Knowledge Distillation in Automated Annotation: \\ Supervised Text Classification with LLM-Generated Training Labels}

\author{Nicholas Pangakis \and Samuel Wolken \\
  University of Pennsylvania \\
  \texttt{\{njpang@sas., sam.wolken@asc.\}upenn.edu}} 

\begin{document}
\maketitle
\begin{abstract}

Computational social science (CSS) practitioners often rely on human-labeled data to fine-tune supervised text classifiers. We assess the potential for researchers to augment or replace human-generated training data with surrogate training labels from generative large language models (LLMs). We introduce a recommended workflow and test this LLM application by replicating 14 classification tasks and measuring performance. We employ a novel corpus of English-language text classification data sets from recent CSS articles in high-impact journals. Because these data sets are stored in password-protected archives, our analyses are less prone to issues of contamination. For each task, we compare supervised classifiers fine-tuned using GPT-4 labels against classifiers fine-tuned with human annotations and against labels from GPT-4 and Mistral-7B with few-shot in-context learning. Our findings indicate that supervised classification models fine-tuned on LLM-generated labels perform comparably to  models fine-tuned with labels from human annotators. Fine-tuning models using LLM-generated labels can be a fast, efficient and cost-effective method of building supervised text classifiers.
\end{abstract}

\section{Introduction}

Supervised text classification often relies on human-labeled text data for training and validation. Computational social science (CSS) researchers frequently use these types of supervised models to classify large quantities of text, ranging from news articles on the internet to government documents \citep{grimmer22,lazer}. Collecting training and validation labels generated by humans for these tasks, however, is expensive, slow, and prone to a variety of errors \citep{grimmer2013, Neuendorf16}.

To address these limitations, prior research suggests utilizing few-shot capabilities of generative large language models (LLMs) to annotate text data instead of human annotators \citep{gilardi2023chatgpt}. Generative LLMs are faster and cheaper than human annotators and do not suffer from common human challenges such as limited attention span or fatigue. While this approach has its limitations and generative LLMs do not excel at all text annotation tasks \citep{pangakis}, prior research illustrates that there are numerous circumstances where generative LLMs can produce high quality text-annotation labels.\footnote{See Appendix A.1 for a longer discussion of automated annotation research in CSS.} 

Although past work suggests LLM few-shot annotation is highly effective, it may be cost prohibitive in many settings. Research with text data often involves classifying millions of documents or text samples. For example, a recent CSS article studies a data set of 6.2 million tweets labeled on four dimensions  \citep{hopkins}, a task that would have cost nearly \$9,000 if using GPT-4 alone.\footnote{Appendix A.2 elaborates on costs with LLM annotation.} Using a knowledge distillation approach \citep{dasgupta-etal-2023-cost, gou_distillation_21, hinton2015distilling}, it may be possible to approximate the performance of a larger ``teacher'' model (e.g., GPT-4 \citep{gpt4}, estimated to have over 1.7T parameters \citep{gptparam}) with much smaller and cheaper task-specific ``student'' models (e.g., BERT Base \citep{bert}, approximately 110 million parameters).

In this paper, we evaluate using generative LLMs to create surrogate labels for fine-tuning downstream supervised classification models. Our approach involves first using a generative LLM to label a subset of text samples and then fine-tuning supervised text classifiers with the LLM-generated labels. Using our outlined approach, we replicate 14 classification tasks from recently published CSS articles. We compare several supervised classifiers (i.e., BERT \citep{bert}, RoBERTa \citep{roberta}, DistilBERT \citep{distilbert}, XLNet \citep{yang2020xlnet}, and Mistral-7B \citep{jiang2023mistral}) fine-tuned on varying quantities of either human-labeled samples or GPT-4-labeled samples. We benchmark the supervised classifiers' performance against GPT-4 and Mistral-7B few-shot labels. In a series of ablation experiments, we also explore whether GPT-4 outputs change over time and how well the student models handle noise in the GPT-generated text labels.

A small number of studies have utilized similar approaches in related domains. \citet{chen} use ChatGPT annotations to train various Graph Neural Networks for a fraction of the cost of human annotations. 
\citet{Golde} also harness ChatGPT to create surrogate text data that aligns with a specific valence (i.e., positive and negative) and then subsequently fine-tune a supervised classifier using the synthetic text. Most analogous to our approach here, \citet{wang} train RoBERTa \citep{roberta} and PEGASUS \citep{pegasus} models on labels generated by GPT-3. Despite strong performance across their analyses, \citet{wang}, as well as the previously mentioned studies, exclusively evaluate closed-source models (i.e., GPT-3 and ChatGPT) on popular, publicly available NLP benchmark tasks (e.g., AGNews, DBPedia, etc), which are plausibly included in the training data for the generative LLM. As a result, these analyses cannot offer a clear indication of performance because their results plausibly suffer from contamination \citep{balepur2024artifacts, li,magar, srivastava2024functional}. Put otherwise, strong performance may reflect memorization, which casts doubt on the generalizability of the findings.

To compare supervised classifiers fine-tuned using LLM-generated labels against those fine-tuned with labels from human annotators, researchers must assess performance on tasks less likely to be affected by contamination. To this end,  all 14 of the classification tasks we replicate are conducted on labeled data sets stored in password-protected archives. Each of the classification tasks in our corpus are real CSS applications and contain human-labeled ground-truth annotations.\footnote{Table \ref{tab:articles} and Table \ref{tab:tasks} include a full list of the data sets and classification tasks.}

Our main contributions are as follows:
\begin{enumerate}[noitemsep, topsep=-0.5\baselineskip]
\item Across 14 classifications tasks, supervised models fine-tuned with GPT-generated labels perform comparably to models fine-tuned with human-labeled data.  The median F1 performance gap between models fine-tuned using GPT-labels and models fine-tuned on human-labeled data is only 0.039. While supervised classifiers fine-tuned with LLM-generated labels perform slightly worse than classifiers fine-tuned with human labels, LLM-generated labels can be a fast, efficient and cost-effective method to fine-tune supervised text classifiers.
\item Supervised models fine-tuned on GPT-4 generated labels perform remarkably close to GPT few-shot models, with a median F1 difference of only 0.006 across the classification tasks. 
\item GPT-4 few-shot models and supervised classifiers fine-tuned on GPT-4 generated labels perform significantly better than all other models on \textit{recall}, but noticeably worse on \textit{precision}.
\end{enumerate}

\section{Methodology}

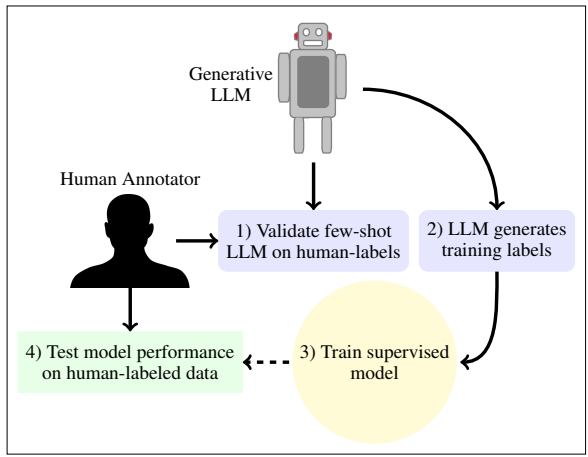
\begin{figure}
    \centering
    \resizebox{\columnwidth}{!}{
    \begin{tikzpicture}[framed]

    \node[inner sep=0pt] (human) at (-3,0)
        {\includegraphics[width=.15\textwidth]{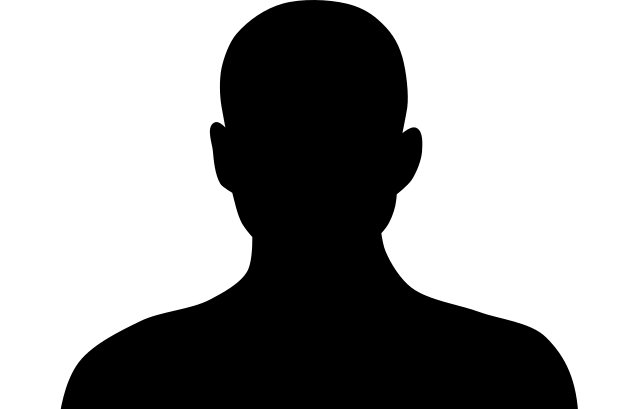}};
    \node[below, align=center, font=\small] at (-3, 1.27) {Human Annotator};
    
    \node[inner sep=0pt] (robot) at (0,2.5)
        {\includegraphics[width=.1\textwidth]{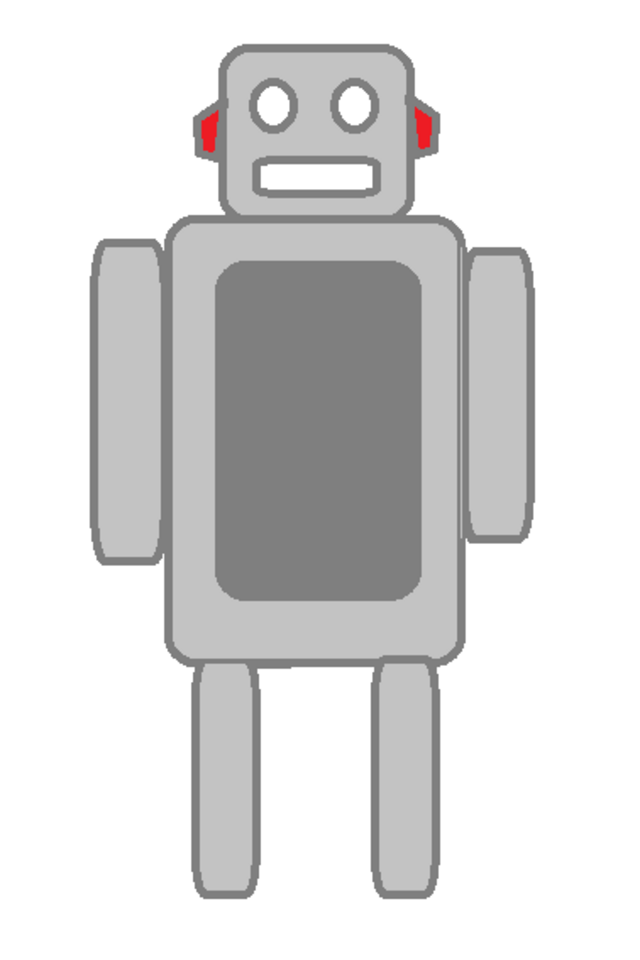}};
        \node[below, align=center, font=\small] at (-1.35, 3) {Generative \\ LLM};

    \node[draw, draw=blue!10, minimum width=2cm, minimum height=1cm,fill=blue!10,font=\small,align=center,rounded corners] (box1) at (0, 0) {1) Validate few-shot \\ LLM on human-labels};
    \node[draw,draw=blue!10, minimum width=2cm, minimum height=1cm, fill=blue!10,font=\small,align=center, rounded corners] (box2) at (3, 0) {2) LLM generates\\training labels};
    
    \draw[fill=yellow!20,draw=yellow!20, align=center,font=\small](1,-2) circle (38 pt) node{3) Train supervised\\model};
    
    \node[draw, draw=green!10,minimum width=2cm, minimum height=1cm,fill=green!10,align=center,font=\small] (box4) at (-3, -2) {4) Test model performance\\on human-labeled data};
    
    % Lines connecting the nodes

    \draw[->, ultra thick] (-2.25, 0) -- (box1.west);
    \draw[->, ultra thick] (human.south) -- (box4.north);
    \draw[->, ultra thick] (robot.south) -- (box1.north);
    \draw[->, ultra thick] (box2.south) to[out=-90, in=0] (2.4, -2);
 
    \draw[->, ultra thick] (robot.east) to[out=0, in=90] (box2.north);
    \draw[->, dashed, ultra thick] (-.4,-2) -- (box4.east);
\end{tikzpicture}}
    \caption{Supervised text classification with LLM-generated training labels.}
    \label{fig:enter-label}
\end{figure}

\begin{figure*}
    \centering
    \includegraphics[width=14.5cm]{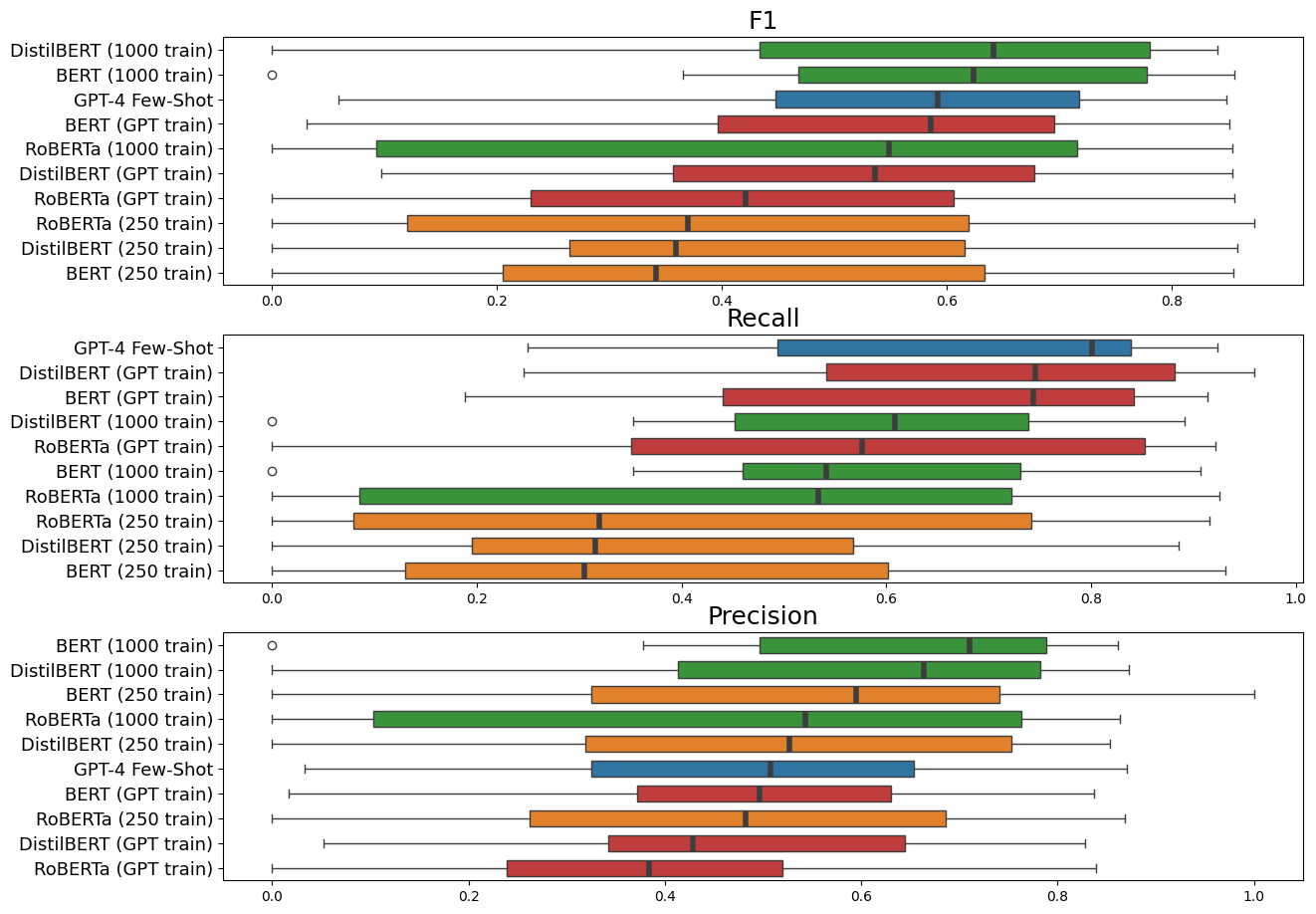}
    \caption{Box plots of performance on test data across 14 tasks. Thick vertical line denotes median. Color represents model type, with green corresponding to models fine-tuned on 1,000 human labels, orange to 250 human labels, red to 1,000 GPT labels, and blue to a few-shot model.} 
    \label{fig1}
\end{figure*}

Figure \ref{fig:enter-label} shows our four-step workflow. First, we validate LLM few-shot performance against a small subset (n=250) of human-labeled text samples for each task. We provide GPT-4\footnote{We select GPT-4 as our main generative model due to its high performance on popular leaderboard websites. In Appendix E.1, we also explore few-shot performance of an open-source model (i.e., Mistral-7B).} with detailed instructions to label the text samples into conceptual categories outlined in the original study.\footnote{We include all prompt details in the supplementary material. We also include our code to query the GPT-4 API.} Because LLM few-shot annotation performance varies across tasks and data sets, validation is always necessary \citep{pangakis}. As such, we validate each generative LLM on a subsample and then adjust the prompt to optimize performance on this initial sample. This process is discussed in greater detail in Appendix C.1. Using the validated prompt, the second step in our workflow involves labeling an additional 1,000 text samples per task using the same generative LLM, which will later be used as data to fine-tune the supervised classifier. 

In the third and fourth steps, we fine-tune a variety of supervised text classifiers and assess performance against a held-out set of 1000 human-labeled samples. Our supervised models include a variety of BERT-family models (i.e., BERT, RoBERTa, and DistilBERT).\footnote{We select these models because of their low cost, speed, and their frequent application in CSS \citep{poli_baseline2, poli_baseline}.} In Appendix E.1, we conduct ablation experiments with XLNet and Mistral-7B. Appendix C.2 describes on our hyperparameter tuning process and additional evaluation details, including how  multi-class tasks were split into separate binary tasks. Ultimately, we compare performance between text classifiers fine-tuned on 1000 LLM-generated samples, 250 human-labeled samples, and 1000 human-labeled samples. 

In addition to analyzing performance across different model architectures and training sample sizes, we also implement a variety of ablation experiments to assess how robust the analyses are to several sources of variance. First, we examine how robust these models are to noisy GPT-generated labels. Specifically, in Appendix E, we implement a novel technique designed to measure noise in GPT-generated labels and then compare supervised models fine-tuned on GPT-generated labels \textit{with noise} against models fine-tuned on GPT-generated labels \textit{without noise}. In a second set of ablation experiments, we replicate the GPT-4 few-shot labels at different points in time. To account for the potential of changing model weights in GPT-4, we re-analyzed each task six months after our initial analyses and compared results across time. Extended discussion and the results for these ablation experiments are shown in Appendix E.

\begin{table*}[t]
    \centering
\begin{tabular}{llcccc}
  \toprule
 Model & Training data & Accuracy & F1 & Precision & Recall \\ 
  \midrule
    GPT-4 & Few shot & 0.88 & 0.59 & 0.51 & 0.80 \\ 
\midrule
\multirow{3}{*}{BERT}  & Human annotation: 250 & 0.89 & 0.34 & 0.59 & 0.30 \\
 & Human annotation: 1000 & \cellcolor{yellow!25}0.92 & \cellcolor{yellow!25} 0.62 & \cellcolor{yellow!25}0.71 & 0.54 \\ 
& GPT-4 annotation: 1000 & 0.87 & 0.59 & 0.50 & \cellcolor{yellow!25}0.74 \\ 
\midrule
\multirow{3}{*}{DistilBERT} & Human annotation: 250 &\cellcolor{yellow!25} 0.89 & 0.36 & 0.53 & 0.32 \\ 
 & Human annotation: 1000 & \cellcolor{yellow!25}0.89 & \cellcolor{yellow!25} 0.64 & \cellcolor{yellow!25}0.66 & 0.61 \\ 
& GPT-4 annotation: 1000 & 0.85 & 0.54 & 0.43 & \cellcolor{yellow!25}0.75 \\ 
\midrule

\multirow{3}{*}{RoBERTa} & Human annotation: 250 & 0.88 & 0.37 & 0.48 & 0.32 \\ 
 & Human annotation: 1000 &\cellcolor{yellow!25} 0.90 & \cellcolor{yellow!25} 0.55 & \cellcolor{yellow!25}0.54 & 0.53 \\
& GPT-4 annotation: 1000 & 0.84 & 0.42 & 0.38 & \cellcolor{yellow!25}0.58 \\
   \bottomrule
\end{tabular}
    \caption{Comparison of classification performance on held-out validation data. Median performance across 14 tasks shown.}
    \label{table1}
\end{table*}

\section{Results}

Classification results for the BERT-family models and GPT-4 few-shot are shown in Table \ref{table1}.\footnote{We conduct few-shot classification by using the classification instructions from the original study as a prompt for the LLM.} In Figure \ref{fig1}, each box plot displays the range of evaluation metrics across all 14 tasks for a given model/training data combination. The thick vertical line denotes the median performance metric across all analyzed tasks. Across all 14 classification tasks, DistilBERT and BERT fine-tuned on 1000 human-samples are the highest performing models, with a median F1 score of 0.641 and 0.624, respectively.\footnote{We use F1 as our primary evaluation criteria due to class imbalance. Full results are shown in Table \ref{tab:complete_results}.} Not far behind, however, is the GPT-4 few-shot model (0.592 median F1) and BERT fine-tuned on 1000 GPT-labeled samples (0.586 median F1). From this we draw two conclusions: First, models fine-tuned on few-shot surrogate labels from a generative LLM perform comparably to models fine-tuned on human labels. Despite a small performance gap, training supervised models on LLM-labeled data can be a quick, effective, and budget-friendly approach for constructing supervised text classifiers. 

Second, models trained on surrogate labels from GPT-4 demonstrate very similar validation performance as labels from GPT-4 with few-shot in-context learning. As each additional GPT-4 query incurs more expense, researchers can save resources by avoiding classifying an entire data set using a generative LLM and instead use them to create training labels for a supervised model.

A secondary finding is that GPT few-shot models and supervised models trained on GPT-generated labels produce remarkably high performance on recall.\footnote{Appendix D displays PR curves for each of the BERT-family supervised models.} GPT-4 few-shot (0.8 median recall) as well as DistilBERT and BERT fine-tuned on GPT-labels (both with 0.746 median recall) achieve significantly better median recall than any model fine-tuned with human labels. The opposite is true for precision: BERT fine-tuned on human-labels achieved the highest precision of the models tested, which was 0.214 higher than median precision for BERT models fine-tuned on GPT-4 labels. Therefore, using surrogate training labels may be better suited for tasks where recall is prioritized over precision.

\section{Discussion}

Surrogate labels from generative LLMs offer a viable, low-resource strategy for fine-tuning task-specific supervised classifiers, but a few points of caution are worth emphasizing. As the variation in our few-shot results indicates, there are  cases where GPT-4 performs poorly on classification tasks. While advancements in LLM technology and additional prompt engineering could mitigate these concerns, it is essential that researchers validate generative LLM performance against ground-truth human-labeled data. Downstream supervised classifiers will not mitigate bias or poor performance in LLM few-shot labels. Thus, while generative LLMs can improve the  classification workflow, their application must remain human-centered.

\clearpage

\section{Limitations}

Here, we identify three main limitations of our analysis. First, as discussed in Section 4 and shown in full detail in Table \ref{tab:complete_results}, there are various circumstances where supervised models fine-tuned on LLM-generated labels fail to produce satisfactory results. This may be due to inaccurate annotations from GPT-4, poor performance from the supervised classifier, or both. While it is possible that additional prompt engineering or hyperparameter tuning could improve performance, it is essential to stress that each of these optimization strategies rely on human labels for comparison. As a result, we argue that it is essential to center human judgement as ground truth when optimizing models and adjudicating between models.

A second, related limitation refers to understanding the errors in the model outputs. Specifically, it is possible that errors from a GPT-trained model produces correlated but unobservable errors. Building a supervised classifier on top of GPT-4 labels would magnify, rather than offset, any such biases. This, too, underscores the importance of human validation and error analysis. It is, of course, also essential to minimize bias by human annotators. For instance, recruiting human annotators from varying demographic backgrounds when conducting an annotation project may diminish the potential for correlated errors across annotators.

Finally, treating human labels as ground truth is an additional limitation. Although most data sets in our analysis employed multiple human coders, it is of course possible that these annotators made correlated errors. As a result, some disagreements between human ground truth labels and surrogate GPT-4 labels may stem from human error. Such errors could bias performance metrics downward for any of the models assessed. Because our primary interest is making comparisons across models, however, we are mainly interested in their relative performance. Because each model would suffer from the same errors in the human labeled data, we do not see this as a significant concern for this analysis. 

For the analysis in this paper, our reliance on text classification tasks and data from peer-reviewed research in high-impact journals helps to mitigate concerns about data annotation quality. The annotation procedures in each of these tasks received IRB approval and was assessed by independent reviewers to be of quality enough for publication in a high-impact journal. Still, it is important to acknowledge that applied researchers should invest in high-quality human labels, even if only to validate generative LLM annotation performance.

\section{Ethics Statement}

Our research complies with the ACL Ethics Policy. Specifically, our research positively contributes to society and human well-being by providing tools that can aid computational social scientists studying the social world. Using the methods we introduce and test will help scientists better understand a wide range of complicated social problems. Because the techniques proposed and assessed in this article require dramatically less resource expenditure than alternatives, our results can help address inequities in resources across researchers.

Due to the inherent risks of deploying biased models, we stress the necessity of human validation throughout our paper. Given the ease and efficiency gains of using generative LLMs to train supervised classifiers, we believe it is essential to build rigorous testing and evaluation standards that are human-centered. This is why we took great efforts to center our analyses on data sets less prone to contamination risks. 

Moreover, our research and data analysis does not cause any harm while also respecting privacy and confidentiality concerns. As we discuss in our data collection procedures in Appendix B, we conformed to each data repository’s usage and replication policies. Each of the original studies received IRB approval and our analyses conformed to the same safety protocols. All collected data was anonymized by the original authors. Appendix C.3 provides additional details on human annotation protocols, which were all conducted by the original studies and received IRB approval.

\bibliography{anthology,custom}

\appendix

\label{sec:appendix}
\setcounter{table}{0}
\renewcommand{\thetable}{A\arabic{table}}
\renewcommand{\thefigure}{A\arabic{figure}}

\section{Appendix: Prior automated annotation research in computational social science}

\begin{table*}[t]
    \centering
\begin{tabular}{c|c|c|c}
  \toprule
 GPT-4: Entire Corpus (n=6.2m) & GPT-4: n=1000 & Crowdworker: n=1000 & Trained Assistant: n=1000  \\ 
  \midrule
   \$8,990 & \$15 & \$124 & \$187  \\ 
   \bottomrule
\end{tabular}
    \caption{Comparing annotation costs applied to Hopkins et al. (2024).}
    \label{compare_costs}
\end{table*}

\subsection{Overview of automated annotation research}

A growing body of research studying automated annotation claims that few-shot classifications from generative LLMs can match humans on annotation tasks \citep{chiang2023can, ding2022gpt3, gilardi2023chatgpt, he2023annollm, mellon22, pan, Rytting, thapa, tornberg, zhu2023chatgpt,ziems23}. For example, \citet{gilardi2023chatgpt} find that LLMs outperform typical crowdsourced human annotators: “[t]he evidence is consistent across different types of texts and time periods. It strongly suggests that ChatGPT may already be a superior approach compared to crowd annotations on platforms such as MTurk.” Analyzing a range of social science applications, \citet{Rytting} similarly write, “GPT-3 can match the performance of human coders [and in] some cases, it even outperforms humans in increasing intercoder agreement scores." \citet{tornberg} argues that automated annotations by LLMs in his analyses are even as accurate as annotations by human experts. While there are clearly circumstances where automated annotation fails to accurately reflect human judgment \citep{unreliable, reiss23}, researchers can safely use automated annotation procedures as long as they validate against human labels not prone to contamination \citep{pangakis}.

\subsection{Costs associated with implementing automated annotation}

While prior research demonstrates that automated annotation can align with human reasoning in many scenarios, directly using the strategies introduced in prior studies to label an entire text corpus would be cost prohibitive when applied to a typical CSS data set, which often contain millions of observations. Consider the cost for using GPT-4 to label a data set of 6.2 million tweets, which is what \citet{hopkins} analyze. At the time of writing, GPT-4 costs \$0.01 per 1k input tokens and \$0.03 per 1k output tokens, with 1000 tokens corresponding to roughly 750 words.\footnote{See \url{https://openai.com/pricing}} The prompt instructions to replicate Hopkins et al. (2024) contained approximately 500 words and the average tweet length was around 25 words. Because the full corpus contained 6.2 million tweets and the code to query the OpenAI API was implemented in batches of 10 tweets, a full automated annotation to process the corpus in \citet{hopkins} would require 620,000 batches fed into GPT-4. Each batch (i.e., 750 words per input) corresponds to roughly 1,000 input tokens, per OpenAI’s suggested benchmark. Since the outputs were standardized, the outputs for these analyses tended to be around 150 tokens. 

Thus, when broken down into tokens, the total number of processed input tokens for this analysis would be \(1,000\times620,000\) and the total processed output tokens would be \(150\times620,000\). When factoring the cost per token for input and output tokens, the total cost comes to \( \$8,990 = (1,000\times620,000\times0.00001) + (150\times620,000\times0.00003)\). While this is a loose estimate, it illustrates the challenges posed by the marginal per-sample cost of automated LLM annotation for large-N CSS research. Using our approach, labeling 1,000 text samples and training a supervised classifier would cost under \$15.

Implementing our proposed workflow also reduces annotation labor costs. For example, hiring crowd-source workers to label a subset of text samples to serve as training observations would still cost significantly more than using automated annotation. \citet{hopkins}, for example, hire MTurk workers and paid them \$0.06 to \$0.07 per task depending on the total number of annotations (\$15.00 per hour for six tasks per minute), which extrapolates to 360 tasks per hour. Under the standard assumption of three MTurk workers per task and taking a majority vote, the entire annotation time to label 1,000 tweets would have taken slightly under three hours and cost \$124. However, due to serious data quality concerns about crowd-workers \citep{Chmielewski20, Douglas23,veselovsky}, a better cost comparison is against trained research assistants instead. Assuming 45 seconds per task and a \$15 hourly rate, manually annotating 1,000 text samples would take 12.5 hours and cost approximately \$187. 

Table \ref{compare_costs} shows a comparison of these costs. Not only is automated annotation remarkably faster than human annotators, our procedures introduced here can cost researchers less than 10\% the cost of typical alternatives. These efficiency gains are conservative in the sense that they disregard the time to find, hire, and train the annotator.\footnote{It is worth stressing here that validation against human-created labels is still essential. Therefore, researchers may want to prioritize their budgets for hiring domain experts to code a small subset of data to serve as validation and test data, as we demonstrate in Figure \ref{fig:enter-label}. Our cost efficiency calculations are based on training data, not validation and test sets.}

\section{Appendix: Data sets}

In this section, we elaborate on the data sets used in our analysis. Our corpus includes 14 classification tasks across five data sets representing recent applications in computational social science. To avoid the potential for contamination, we rely exclusively on data sets stored in password-protected data archives (e.g., Dataverse). We draw from research published in outlets across a spectrum of disciplines ranging from interdisciplinary publications (e.g., \textit{Proceedings of the National Academy of Sciences}) to high-impact field journals in social science (e.g., \textit{American Journal of Political Science}). To find these articles, we searched journals for articles related to computational social science that implemented some type of manual annotation procedure. The human-labeled data from the original study is treated as the ground truth. We discuss the human annotation procedures in the original studies at greater length in Appendix C.3.

It is important to note that while the raw data (e.g., tweets and Facebook posts) may be included in the LLM pretraining data, the accompanying labels from the human annotators are certainly not included in the pretraining data. This is because the labels accompanying each text sample (e.g., whether a tweet referenced a specific racial identity frame) are not public-facing. If the text without the associated label is  not included in the pretraining data, there is no cause for concern that the annotation task would suffer from contamination. 

Table \ref{tab:articles} and Table \ref{tab:tasks} contain the full details for every task and data set. Overall, our data encompass diverse degrees of class imbalance: Across tasks, the mean positive class frequency is 16.2\%, the minimum is 0.04\%, and the maximum is 61\%. The sources of labels are representative of common approaches to annotation: 42.9\% of tasks were annotated by crowdsourced workers, 28.6\% by experts, and 28.6\% by research assistants.

Our replications involve fine-tuning supervised classifiers using manually annotated data from the replication data sets. For every replication classification task, we conformed to each data repository’s replication policies. Each of the original studies received IRB approval and our analyses conformed to the same safety protocols, including full anonymization and agreeing to not publicly post the raw data without permission. As such, our replication of each data set is compatible with its intended usage. 

Although all of the data sets were anonymized before our replications, we manually reviewed each data set to confirm privacy protections. One of the data sets \citep{Saha23} contains hate speech, but this is because it is a central part of the research question from the original study. As a result, we include examples of hate speech in that particular replication. From manual review, no other data set contained offensive material.

\nocite{card23, Saha23, muller, peng}

\begin{table*}[b] 
    \centering
    \begin{tabular}{p{3cm}p{7cm}p{4cm}p{1cm}}
         \toprule
         Author(s) & Title & Journal & Year \\
         \midrule
         Card et al. & Computational analysis of 140 years of US political speeches reveals more positive but increasingly polarized framing of immigration & PNAS & 2022 \\
    
         \midrule
         Hopkins, Lelkes, and Wolken & The Rise of and Demand for Identity-Oriented Media Coverage & American Journal of Political Science & 2024 \\
         \midrule
         Müller & The Temporal Focus of Campaign Communication & Journal of Politics & 2021 \\
         \midrule
         Peng, Romero, and Horvat & Dynamics of cross-platform attention to retracted papers & PNAS & 2022 \\
         \midrule
         Saha  et al. & On the rise of fear speech in online social media  & PNAS & 2022 \\
         \bottomrule
         & 
    \end{tabular}
    \caption{Replication data sources.}
    \label{tab:articles}
\end{table*}

\begin{table*}[b]
\centering
\begin{tabular}{p{3cm}p{1cm}p{1.5cm}p{8cm}}
\toprule
\multicolumn{1}{c}{Study} & \multicolumn{1}{c}{\# of tasks} & \multicolumn{1}{c}{Annotation source}  & \multicolumn{1}{c}{Classification tasks}                                       
\\ \midrule
Card et al. (2022) & 4 & Research assistants &  Classify US congressional speeches to identify whether the speech discussed immigration or immigration policy, along with an accompanying tone: pro-immigration, anti-immigration, or neutral.                      
\\ \midrule
Hopkins, Lelkes, and Wolken (2024) & 4 & Crowd & Classify headlines, Tweets, and Facebook share blurbs to identify references to social groups defined by a) race/ethnicity; b) gender/sexuality; c) politics; d) religion. 
\\ \midrule
Müller (2021)  & 3  & Expert & Classify sentences from political party manifestos for temporal direction: past, present, or future.                                                                                                                                                          \\ \midrule
Peng, Romero, and Horvat (2022) & 1    & Expert & Classify whether Tweets express criticism of findings from academic papers.                                                                                                                   \\ \midrule
Saha et al. (2020) & 2     &    Crowd        & Classify social media posts into fear speech, hate speech, both, or neither.                      
                \\ \bottomrule
\end{tabular}
\caption{Descriptions of replication classification tasks.}
\label{tab:tasks}
\end{table*}

\begin{table*}[b]
\centering
\resizebox{\textwidth}{!}{
\begin{tabular}{lll|cccc|cccc|cccc|cccc}
\toprule
\multirow{4}{*}{Data set}  & \multirow{4}{*}{Task}  & \multirow{4}{*}{Model}  & \multicolumn{16}{c}{\textbf{Training data}}  \\ 

  &  & & \multicolumn{4}{c}{Few shot}
 & \multicolumn{4}{c}{Human: 250}
 & \multicolumn{4}{c}{Human: 1000}
 & \multicolumn{4}{c}{GPT: 1000} \\ 

  \cmidrule(lr){4-7} \cmidrule(lr){8-11} \cmidrule(lr){12-15}\cmidrule(lr){16-19}
  & & & 
   Ac. & F1 & Pr. & Re. & Ac. & F1 & Pr. & Re. & Ac. & F1 & Pr. & Re. & Ac. & F1 & Pr. & Re.
%\multicolumn{4}{c}{\begin{rotate}{60} Few shot \end{rotate}}
% & \multicolumn{4}{c}{\begin{rotate}{60} Human: 250 \end{rotate}}
% & \multicolumn{4}{c}{\begin{rotate}{60} Human: 1000 \end{rotate}}
% & \multicolumn{4}{c}{\begin{rotate}{60} GPT: 250 \end{rotate}} 
  \\
  \midrule
\multirow{16}{*}{Card et al.} & \multirow{4}{*}{Cat: Neg} & GPT-4 & 0.85 & 0.65 & 0.54 & 0.83 &  &  &  &  &  &  &  &  &  &  &  &  \\ 
   &  & BERT &  &  &  &  & 0.88 & 0.58 & 0.74 & 0.48 & 0.87 & 0.56 & 0.65 & 0.49 & 0.81 & 0.56 & 0.47 & 0.72 \\ 
   & & RoBERTa &  &  &  &  & 0.85 & 0.51 & 0.59 & 0.45 & 0.84 & 0.48 & 0.55 & 0.42 & 0.78 & 0.57 & 0.43 & 0.82 \\ 
   &  & DistilBERT &  &  &  &  & 0.86 & 0.56 & 0.61 & 0.51 & 0.86 & 0.58 & 0.61 & 0.55 & 0.81 & 0.58 & 0.47 & 0.74 \\ 
   \cmidrule(lr){4-19} 
   & \multirow{4}{*}{Cat: Imm} & GPT-4 & 0.81 & 0.81 & 0.74 & 0.90 &  &  &  &  &  &  &  &  &  &  &  &  \\ 
   &  & BERT &  &  &  &  & 0.85 & 0.84 & 0.79 & 0.89 & 0.86 & 0.86 & 0.81 & 0.91 & 0.84 & 0.83 & 0.76 & 0.91 \\ 
   & & RoBERTa &  &  &  &  & 0.86 & 0.85 & 0.80 & 0.92 & 0.85 & 0.84 & 0.77 & 0.92 & 0.82 & 0.82 & 0.74 & 0.92 \\ 
  & & DistilBERT &  &  &  &  & 0.85 & 0.84 & 0.80 & 0.88 & 0.84 & 0.84 & 0.79 & 0.89 & 0.82 & 0.82 & 0.73 & 0.92 \\ 
  \cmidrule(lr){4-19} 
  & \multirow{4}{*}{Cat: Neut.} & GPT-4  & 0.83 & 0.26 & 0.27 & 0.25 &  &  &  &  &  &  &  &  &  &  &  &  \\ 
   &  & BERT &  &  &  &  & 0.80 & 0.35 & 0.29 & 0.44 & 0.85 & 0.36 & 0.38 & 0.35 & 0.87 & 0.38 & 0.44 & 0.34 \\ 
  &  & RoBERTa &  &  &  &  & 0.88 & 0.30 & 0.46 & 0.23 & 0.88 & 0.00 & 0.00 & 0.00 & 0.84 & 0.33 & 0.33 & 0.34 \\ 
  &  & DistilBERT &  &  &  &  & 0.85 & 0.28 & 0.32 & 0.25 & 0.85 & 0.36 & 0.37 & 0.35 & 0.86 & 0.38 & 0.40 & 0.36 \\ 
  \cmidrule(lr){4-19} 
  & \multirow{4}{*}{Cat: Pro}& GPT-4 & 0.88 & 0.50 & 0.55 & 0.46 &  &  &  &  &  &  &  &  &  &  &  &  \\ 
   &  & BERT &  &  &  &  & 0.86 & 0.33 & 0.44 & 0.27 & 0.84 & 0.44 & 0.42 & 0.46 & 0.87 & 0.45 & 0.51 & 0.40 \\ 
   &  & RoBERTa &  &  &  &  & 0.87 & 0.37 & 0.51 & 0.30 & 0.84 & 0.37 & 0.41 & 0.34 & 0.85 & 0.41 & 0.43 & 0.39 \\ 
 &  & DistilBERT &  &  &  &  & 0.87 & 0.29 & 0.55 & 0.19 & 0.83 & 0.38 & 0.38 & 0.37 & 0.84 & 0.35 & 0.40 & 0.31 \\ 
 \midrule
  \multirow{16}{*}{Hopkins et al.} & \multirow{4}{*}{Political} & GPT-4 & 0.88 & 0.43 & 0.30 & 0.79 &  &  &  &  &  &  &  &  &  &  &  &  \\ 
  &  & BERT &  &  &  &  & 0.95 & 0.32 & 0.60 & 0.22 & 0.96 & 0.62 & 0.71 & 0.54 & 0.82 & 0.34 & 0.21 & 0.82 \\ 
   &  & RoBERTa &  &  &  &  & 0.84 & 0.37 & 0.23 & 0.85 & 0.96 & 0.62 & 0.73 & 0.54 & 0.84 & 0.37 & 0.23 & 0.85 \\ 
   &  & DistilBERT &  &  &  &  & 0.94 & 0.29 & 0.50 & 0.20 & 0.96 & 0.63 & 0.72 & 0.56 & 0.83 & 0.34 & 0.22 & 0.80 \\ 
   \cmidrule(lr){4-19} 
  & \multirow{4}{*}{Gender} & GPT-4 & 0.95 & 0.74 & 0.68 & 0.82 &  &  &  &  &  &  &  &  &  &  &  &  \\ 
   &  & BERT &  &  &  &  & 0.91 & 0.20 & 0.46 & 0.13 & 0.96 & 0.80 & 0.86 & 0.74 & 0.94 & 0.72 & 0.62 & 0.85 \\ 
  &  & RoBERTa &  &  &  &  & 0.91 & 0.08 & 0.44 & 0.04 & 0.95 & 0.73 & 0.78 & 0.68 & 0.92 & 0.67 & 0.54 & 0.87 \\ 
  &  & DistilBERT &  &  &  &  & 0.94 & 0.52 & 0.83 & 0.38 & 0.97 & 0.81 & 0.87 & 0.75 & 0.93 & 0.71 & 0.59 & 0.88 \\ 
  \cmidrule(lr){4-19} 
   & \multirow{4}{*}{Race} & GPT-4 & 0.96 & 0.57 & 0.41 & 0.92 &  &  &  &  &  &  &  &  &  &  &  &  \\ 
  &  & BERT &  &  &  &  & 0.97 & 0.00 & 0.00 & 0.00 & 0.98 & 0.56 & 0.71 & 0.46 & 0.98 & 0.64 & 0.54 & 0.77 \\ 
   &  & RoBERTa &  &  &  &  & 0.97 & 0.00 & 0.00 & 0.00 & 0.97 & 0.00 & 0.00 & 0.00 & 0.97 & 0.59 & 0.45 & 0.85 \\ 
   &  & DistilBERT &  &  &  &  & 0.97 & 0.00 & 0.00 & 0.00 & 0.99 & 0.71 & 0.77 & 0.65 & 0.97 & 0.54 & 0.46 & 0.65 \\ 
   \cmidrule(lr){4-19} 
  & \multirow{4}{*}{Religion} & GPT-4 & 0.98 & 0.61 & 0.47 & 0.88 &  &  &  &  &  &  &  &  &  &  &  &  \\ 
   &  & BERT &  &  &  &  & 0.98 & 0.21 & 1.00 & 0.12 & 0.99 & 0.73 & 0.75 & 0.71 & 0.98 & 0.61 & 0.48 & 0.82 \\ 
   &  & RoBERTa &  &  &  &  & 0.98 & 0.00 & 0.00 & 0.00 & 0.98 & 0.00 & 0.00 & 0.00 & 0.98 & 0.00 & 0.00 & 0.00 \\ 
   &  & DistilBERT &  &  &  &  & 0.98 & 0.00 & 0.00 & 0.00 & 0.99 & 0.69 & 0.67 & 0.71 & 0.97 & 0.53 & 0.37 & 0.94 \\ 
   \midrule
  \multirow{12}{*}{Müller} &  \multirow{4}{*}{Future} & GPT-4 & 0.82 & 0.85 & 0.87 & 0.83 &  &  &  &  &  &  &  &  &  &  &  &  \\ 
   &  & BERT &  &  &  &  & 0.83 & 0.85 & 0.88 & 0.84 & 0.82 & 0.85 & 0.85 & 0.85 & 0.81 & 0.85 & 0.84 & 0.87 \\ 
   &  & RoBERTa &  &  &  &  & 0.84 & 0.87 & 0.87 & 0.88 & 0.82 & 0.85 & 0.86 & 0.85 & 0.82 & 0.86 & 0.84 & 0.87 \\ 
   &  & DistilBERT &  &  &  &  & 0.83 & 0.86 & 0.85 & 0.86 & 0.81 & 0.84 & 0.87 & 0.82 & 0.82 & 0.85 & 0.83 & 0.88 \\ 
   \cmidrule(lr){4-19} 
   & \multirow{4}{*}{Past} & GPT-4  & 0.91 & 0.74 & 0.66 & 0.84 &  &  &  &  &  &  &  &  &  &  &  &  \\ 
  &  & BERT &  &  &  &  & 0.94 & 0.83 & 0.74 & 0.93 & 0.95 & 0.83 & 0.80 & 0.85 & 0.93 & 0.79 & 0.71 & 0.89 \\ 
   &  & RoBERTa &  &  &  &  & 0.94 & 0.80 & 0.81 & 0.79 & 0.95 & 0.85 & 0.79 & 0.92 & 0.85 & 0.00 & 0.00 & 0.00 \\ 
   &  & DistilBERT &  &  &  &  & 0.94 & 0.79 & 0.77 & 0.80 & 0.94 & 0.80 & 0.79 & 0.82 & 0.93 & 0.79 & 0.68 & 0.96 \\ 
   \cmidrule(lr){4-19} 
   & \multirow{4}{*}{Present} & GPT-4 & 0.82 & 0.62 & 0.64 & 0.60 &  &  &  &  &  &  &  &  &  &  &  &  \\ 
   &  & BERT &  &  &  &  & 0.83 & 0.65 & 0.66 & 0.64 & 0.83 & 0.65 & 0.64 & 0.66 & 0.81 & 0.61 & 0.63 & 0.58 \\ 
   &  & RoBERTa &  &  &  &  & 0.84 & 0.66 & 0.71 & 0.61 & 0.84 & 0.68 & 0.68 & 0.67 & 0.83 & 0.61 & 0.68 & 0.56 \\ 
   & & DistilBERT &  &  &  &  & 0.83 & 0.64 & 0.69 & 0.59 & 0.83 & 0.65 & 0.66 & 0.64 & 0.82 & 0.59 & 0.66 & 0.54 \\ 
   \midrule
  \multirow{4}{*}{Peng et al.} & \multirow{4}{*}{Critical} & GPT-4  & 0.85 & 0.54 & 0.48 & 0.63 &  &  &  &  &  &  &  &  &  &  &  &  \\ 
   &  & BERT &  &  &  &  & 0.87 & 0.43 & 0.59 & 0.34 & 0.91 & 0.63 & 0.76 & 0.54 & 0.79 & 0.43 & 0.35 & 0.56 \\ 
  &  & RoBERTa &  &  &  &  & 0.88 & 0.44 & 0.61 & 0.34 & 0.87 & 0.62 & 0.54 & 0.73 & 0.78 & 0.43 & 0.34 & 0.59 \\ 
   &  & DistilBERT &  &  &  &  & 0.83 & 0.43 & 0.42 & 0.44 & 0.86 & 0.54 & 0.50 & 0.58 & 0.77 & 0.41 & 0.33 & 0.56 \\ 
   \midrule
  \multirow{8}{*}{Saha et al.} & \multirow{4}{*}{CV} & GPT-4  & 0.97 & 0.06 & 0.03 & 0.25 &  &  &  &  &  &  &  &  &  &  &  &  \\ 
  &  & BERT &  &  &  &  & 1.00 & 0.00 & 0.00 & 0.00 & 1.00 & 0.00 & 0.00 & 0.00 & 0.94 & 0.03 & 0.02 & 0.25 \\ 
   &  & RoBERTa &  &  &  &  & 1.00 & 0.00 & 0.00 & 0.00 & 1.00 & 0.00 & 0.00 & 0.00 & 0.93 & 0.05 & 0.03 & 0.50 \\ 
   &  & DistilBERT &  &  &  &  & 1.00 & 0.00 & 0.00 & 0.00 & 0.99 & 0.00 & 0.00 & 0.00 & 0.94 & 0.10 & 0.05 & 0.75 \\ 
   \cmidrule(lr){4-19} 
   & \multirow{4}{*}{HD} & GPT-4 & 0.88 & 0.35 & 0.28 & 0.45 &  &  &  &  &  &  &  &  &  &  &  &  \\ 
   &  & BERT &  &  &  &  & 0.91 & 0.17 & 0.24 & 0.13 & 0.92 & 0.41 & 0.45 & 0.38 & 0.90 & 0.21 & 0.24 & 0.19 \\ 
   &  & RoBERTa &  &  &  &  & 0.92 & 0.24 & 0.35 & 0.19 & 0.92 & 0.47 & 0.43 & 0.52 & 0.91 & 0.20 & 0.26 & 0.16 \\ 
   &  & DistilBERT &  &  &  &  & 0.91 & 0.26 & 0.32 & 0.22 & 0.91 & 0.40 & 0.38 & 0.42 & 0.91 & 0.28 & 0.33 & 0.25 \\ 
   \bottomrule
\end{tabular}}
\caption{Complete task-by-task classification performance results. Ac., Pr., and Re. refer to accuracy, precision, and recall, respectively.}
    \label{tab:complete_results}
\end{table*}

\section{Appendix: Additional methodological details}

\subsection{Prompt tuning}

As discussed in Section 2, for every task we adjusted each GPT-4 prompt with a human-in-the-loop update procedure to optimize for accurate annotations. This human-in-the-loop process involved three steps. First, we used the generative LLM to annotate a small subset of the text samples per task (n=250).\footnote{This subset of text samples was not included in the held-out test set.} Second, we manually reviewed instances where humans and the generative LLM disagreed on the text's label. Because our accuracy at this stage hovered around 0.8, this usually entailed manually reviewing roughly 50 text labels. Third, we adjusted the prompt instructions to clarify instances where automated annotation failed to correctly align with human judgment.

\begin{figure*}
    \centering
    \includegraphics[width = .9 \textwidth]{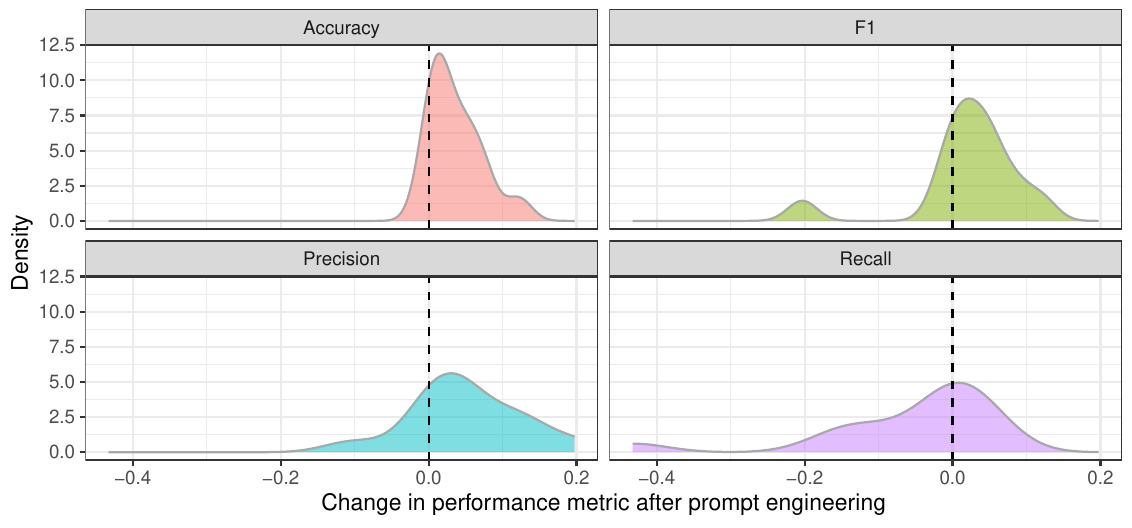}
    \caption{Change in LLM annotation performance on training data after one round of prompt optimization}
    \label{fig:codebook_updates}
\end{figure*}

The prompt tuning process should be minimal (e.g., one or two iterations), because any further efforts could lead to overfitting the prompt to a small subset of the data \citep{egami22}. 
If the prompt is overly tailored to a small subset of the data, then the instructions may not generalize to unseen data. Moreover, if the researcher makes major changes to the prompt, there may be a mismatch between the human annotator’s codebook and the generative LLM’s instructions. Like the previous concern, the differences in the instructions could lead to poor performance on a held-out set. As a result, if there are substantial changes made to the LLM’s prompt, then the researcher should also change the human codebook as well and re-annotate new text samples. As such, these procedures should not be resource or time intensive. Instead, prompt tuning is intended to be a part of a validation process of few-shot in-context learning.

Some researchers argue that small changes to the LLM prompt instructions can dramatically alter automated annotation performance \citep{reiss23}, whereas others claim that alterations have a marginal effect \citep{Rytting}. To test how variations in the prompt instructions affect performance, we evaluated automated annotation performance before and after the prompt tuning process. 

Figure \ref{fig:codebook_updates} shows the distributions of change in performance metrics after updating the LLM prompt and re-annotating the same text samples. This analysis demonstrates whether and how prompt optimization affects LLM annotation, holding constant the data and conceptual categories. In most cases, prompt optimization led to minor improvement in accuracy and F1---although recall decreased in more cases than improved after updating the prompts. The small magnitude of change in classification performance suggests that generative LLMs are fairly robust to slight word changes in the prompt, which aligns with prior work that conducts similar experiments \citep{Rytting}. While the magnitude of improvement was generally small, researchers experiencing subpar LLM annotation performance can use human-in-the-loop prompt optimization to ensure that their instructions are not the cause of poor performance. 

Qualitatively, the most common mistakes we observed by the generative LLM during the prompt optimization stage were false positives stemming from the text sample containing language broadly associated with the conceptual category of interest. For example, one task focused on identifying immigration content in American political speeches \citep{card23}. Initially, the generative model consistently categorized a text sample as containing an immigration reference if the speech mentioned a foreign country or foreign national, irrespective of whether the mention was connected to immigration in any way. For the prompt-update process for this task, changes in this case meant clarifying that any reference to a foreign country or foreign national did not warrant a positive class instance unless it was explicitly referenced in relation to American immigration or immigration policy. While this process was manual, we also believe that future work could conduct these procedures algorithmically—plausibly using generative AI as well.

\subsection{Hyperparameter tuning, evaluation, and compute details}

Our experiments involved varying the training data used to fine-tune numerous supervised classifiers (i.e., 250 human samples, 1000 human samples, and 1000 GPT-labeled samples). To select each supervised classifier, we implemented a grid search over 18 possible hyperparameter combinations. In particular, we optimized learning rate (1e-5, 2e-5, and 5e-5), batch size (8 and 16), and epochs (2, 4, and 6). We conducted our search on a subsample of 250 text samples per task and retained the best hyperparameters (in terms of highest F1) across each task. We subsequently used the best-performing combination of hyperparameters for all applications of a specific model (see best-performing hyperparameter configurations in Table \ref{tab:hyper}). Despite not adopting a more exhaustive approach to hyperparameter tuning, we observe strong performance across our classification tasks, with a few exceptions.  Table \ref{tab:arch} displays additional model hyperparameters that remained constant across tasks, as well as basic information about each model's architecture.

Overall, for each task we had a total of 2,500 labeled text samples labeled by both human annotators and the LLM: (1) a training set of 1,000 text samples; (2) two separate validation sets (both with n=250); and (3) a test set (n=1000). Each of these sets of data were labeled by humans and the generative LLM. The training set (n=1000) was used to fine-tune the supervised classifiers. The first validation set (n=250) was used to optimize the generative LLM prompt and validate its few-shot performance. The second, separate validation set (n=250) was used to conduct our grid search. The test set (n=1000) was used to assess the final performance of the few-shot model and the supervised models.

For all 14 tasks, evaluation was conducted on a test set of 1000 held-out text samples that had previously been labeled by human annotators. To harmonize the diverse range of annotation tasks into a common framework for evaluation, we treat every task dimension as a separate binary annotation task. Thus, if an article included a classification task with three potential labels, we split the annotation process into three discrete binary classification tasks. As is standard in binary classification evaluation, we report accuracy, F1, precision, and recall for every task and model.\footnote{Because our tasks are all binary, there is no need to report any multi-label classification metrics, like Macro-F1.} Table \ref{tab:complete_results} displays the full classification results across all tasks and models. 

All of our supervised training analyses were implemented in Python 3.10.12 with HuggingFace's Transformers \citep{wolf} and PyTorch libraries \citep{Paszke}. We conducted all data preprocessing in Python Pandas \citep{pandas}. Our computing infrastructure was Google Colab, where we used 215 T4 GPU compute units (roughly 421.4 GPU hours). As with our model selection, we chose this computing environment due to its low cost and ease of application. Any computational social scientist could conduct the same analyses. In the supplementary material, we include all code to run our supervised training procedures. 

\begin{table*}[b] 
    \centering
    \begin{tabular}{p{2.5 cm}|p{5.5cm}p{4cm}}
         \toprule
         Study & Task & Hyperparameters \\
         \midrule
            \multirow{4}{*}{Card et al.} & Classify immigration speeches & learning rate (5e-05), batch size (8), epochs (4) \\
            \cmidrule(lr){2-3} 
              & Classify pro-immigration speeches & learning rate (5e-05), batch size (16), epochs (6) \\
              \cmidrule(lr){2-3}
            & Classify anti-immigration speeches & learning rate (5e-05), batch size (8), epochs (6) \\
            \cmidrule(lr){2-3}
             & Classify neutral immigration speeches & learning rate (5e-05), batch size (8), epochs (4) \\
         \midrule
            \multirow{4}{*}{Hopkins et al.} & Classify race/ethnicity & learning rate (5e-05), batch size (8), epochs (4) \\
            \cmidrule(lr){2-3}
              & Classify gender & learning rate (5e-05), batch size (8), epochs (6) \\
         \cmidrule(lr){2-3}
              & Classify political groups & learning rate (5e-05), batch size (16), epochs (6) \\
         \cmidrule(lr){2-3}
             & Classify religious groups & learning rate (5e-05), batch size (8), epochs (6) \\
             \midrule
         \multirow{3}{*}{Müller} & Classify past & learning rate (5e-05), batch size (8), epochs (4) \\
         \cmidrule(lr){2-3}
              & Classify present & learning rate (5e-05), batch size (8), epochs (4) \\
         \cmidrule(lr){2-3}
              & Classify future & learning rate (2e-05), batch size (8), epochs (6) \\
         \midrule
         Peng et al. & Classify criticism & learning rate (5e-05), batch size (8), epochs (6) \\
         \midrule
       \multirow{2}{*}{Saha  et al.} & Classify fear speech & learning rate (5e-05), batch size (8), epochs (6) \\
         \cmidrule(lr){2-3}
              & Classify hate speech & learning rate (5e-05), batch size (8), epochs (4) \\
         \bottomrule
    \end{tabular}
    \caption{Hyperparameter settings per task.}
    \label{tab:hyper}
\end{table*}

\begin{table*}[b] 
    \centering
    \begin{tabular}{p{4 cm}p{1.5 cm}p{1.5cm}p{1.5cm}p{1.5cm}p{1.5cm}}
         \toprule
          & BERT-base & RoBERTa-base & DistilBERT & XLNet-base & Mistral-7B \\
         \midrule
            \# parameters & 110m & 125m & 66m & 110m & 7b \\
         \midrule
         \# attention heads & 12 & 12 & 12 & 12 & 32\\
         \midrule
          Hidden dim. & 768 & 768 & 768 & 768 & 4096 \\
         \midrule
         Feedforward dim. & 3072 & 3072 & 3072 & 3072 & 14336\\
         \bottomrule
         & 
    \end{tabular}
    \caption{Model architectures and additional hyperparameters.}
    \label{tab:arch}
\end{table*}

\subsection{Additional details on human annotation procedures}

We introduce a novel corpus of labeled text data for annotations. To create this data set, we compile labeled data from recent studies, as detailed in \ref{tab:articles}. As a result, we did not work with annotators to generate any original data. We adopted materials from these original studies instead. While we do not report the instructions given to each study's human annotators, we do provide the prompt instructions that were used to query GPT-4 in the supplementary material. These instructions were taken directly from the original study's human annotator instructions. All additional details on the annotation procedures (e.g., how they were recruited, payment, consent, and demographic characteristics) can be found in the original studies' supplementary material. 

While we do not describe each study's procedures in detail, we manually selected our annotation studies due to their high-quality human labeling practices. All of the replicated studies were approved by an IRB. These studies all deployed either expert coders or numerous non-expert coders of varying backgrounds. Because all of the human annotation text is part of the peer-review process in high-impact journals and due to the strict annotation guidelines and principles these studies adhered to, we conclude that the human annotations are of high-quality. 

\section{Appendix: Extended results}

\begin{figure*}
    \centering
    \includegraphics[width=\textwidth]{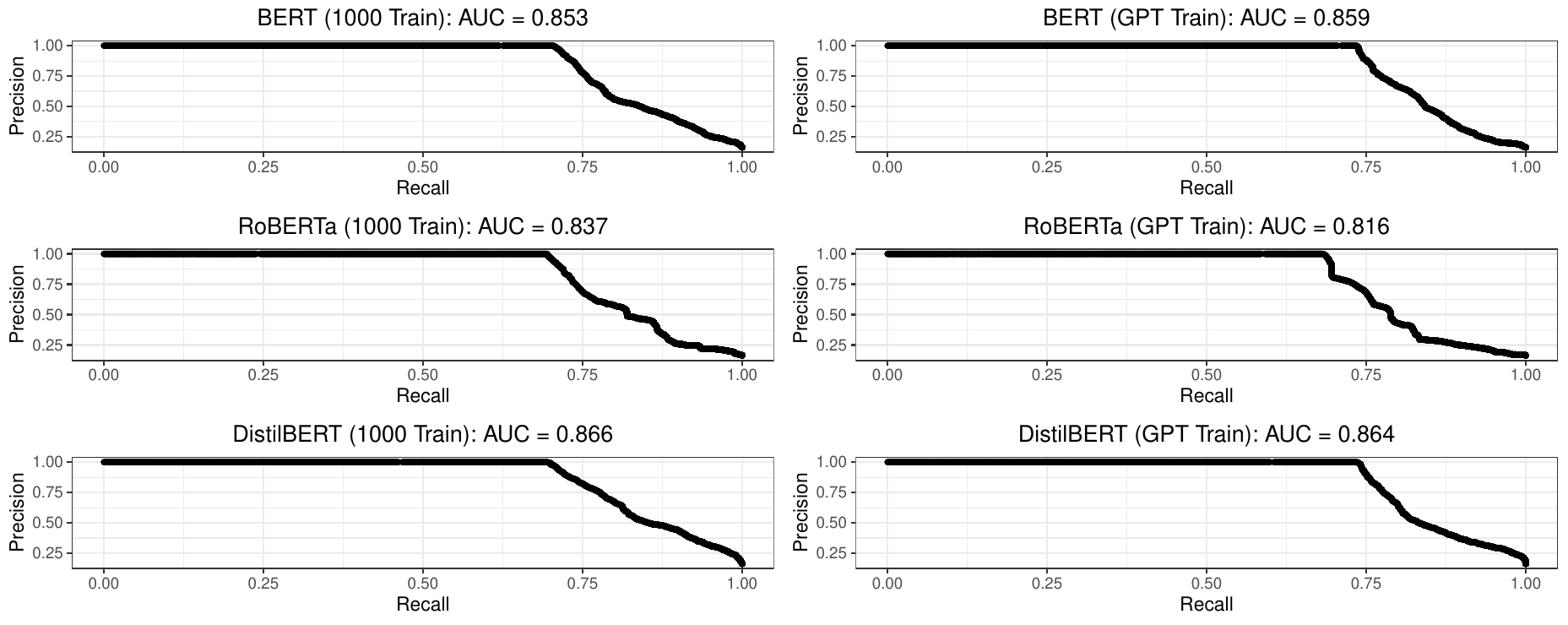}
    \caption{Precision-recall curves across each BERT-family model}. 
    \label{fig-pr}
\end{figure*}

Figure \ref{fig-pr} shows precision-recall (PR) curves for each of the BERT-family models trained on either human labels or GPT labels, pooling all classification tasks. The decrease in performance for GPT-generated labels compared with human labels is small based on area under the curve (AUC). Thus, supervised classifiers trained with GPT-generated labels perform comparably to classifiers trained with human-generated labels on these tasks. Across models and tasks, precision appears to drop below 1.0 around 0.7 recall.

\section{Appendix: Ablation experiments}

We conducted a variety of ablation experiments to account for sources of variance. The next three sections detail these experiments and their main findings.

\subsection{Comparing classifiers with different model size and architecture}

First, to account for variation in model architecture and model size, we compare performance across two additional language models for supervised classification (i.e., XLNet and Mistral-7B). These models are beyond the BERT-family models included in the main analyses (i.e., BERT, DistilBERT, and RoBERTa). In addition to a Mistral-7B supervised sequence classification model, we also generate few-shot labels using Mistral-7B using the same procedures we employed in the GPT-4 few-shot model. 

The primary difference between the BERT-family models and XLNet is the training objective. The BERT-based models are pretrained using a Masked Language Modeling (MLM) objective, whereas XLNet is an autoregressive model that uses Permutation Language Modeling (PLM), which involves learning context across input tokens in any permutation order. In addition to being significantly larger  than the BERT models, Mistral-7B utilizes a distinct type of attention in the pretraining process (i.e., grouped-query attention (GQA) and sliding window attention (SWA)). We include the Mistral-7B few-shot model as a smaller, open-source alternative to GPT-4. Mistral-7B was selected because the model weights are available for download and it displays higher performance than Llama-13B \citep{jiang2023mistral}.

\begin{figure*}
    \centering
    \includegraphics[width=\textwidth]{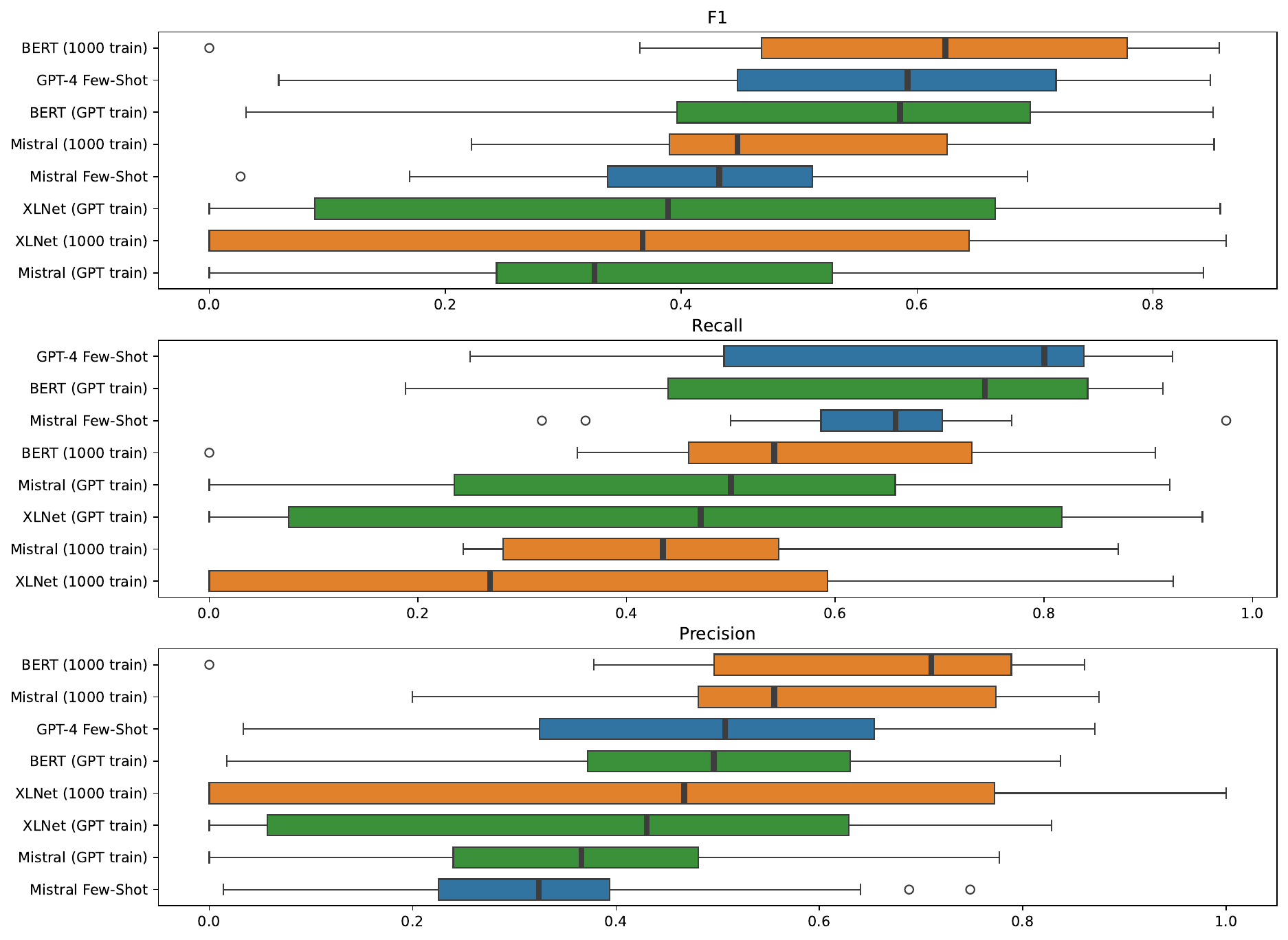}
    \caption{Box plots of ablation performance on test data across 14 tasks. Thick vertical line denotes median.}. 
    \label{fig-abstract}
\end{figure*}

Figure \ref{fig-abstract} shows the classification performance from these additional models and compares them to the results from BERT and GPT-4 few-shot in the main analyses. The test set for these analyses is the same as the main analysis shown in the paper. Our results from examining these additional models do not change the substantive conclusions in the paper:  Models trained on surrogate training labels perform comparably to models trained with human labeled data. XLNet even performs slightly better than the fully human labels. The gap between Mistral-7b fine-tuned using human labels and GPT-labels, however, is notably larger than the other models, with a median difference of 0.12. Overall, BERT and GPT-4 still appear to be the strongest performing models.

There is also a fairly sizeable gap between the open-source (Mistral-7B) and closed-source (GPT-4) few-shot models. Although it may be expected from a significantly smaller and free-to-use model, F1 scores for Mistral-7B are 0.16 worse, on average, than GPT-4. Mistral-7B also took significantly longer to run than GPT-4. These findings further reinforce the necessity of human validation.

\subsection{Comparing classifiers with and without noise}

Our second set of ablation experiments involve comparing supervised models trained on GPT-generated labels \textit{with noise} against GPT-generated labels \textit{without noise}. To measure noise in the GPT-labels, we utilize the predicted token sampling process of generative LLMs to gauge an LLM's ``confidence” in the annotation of each text sample. By introducing randomness in the LLM sampling process through the temperature setting and by repeatedly classifying the same text sample multiple times, we identify text samples that cannot be clearly classified into one of the annotation categories specified by the prompt instructions.\footnote{Generative LLMs output a series of probabilities that correspond to each token in its vocabulary. To select a specific token from this probability distribution, generative LLMs sample a randomly selected token, weighted by its probability. The temperature hyperparameter governs this sampling process. A higher temperature setting flattens the probability distribution and causes the sampling draw to become more uniform across tokens. A lower temperature, however, isolates the sampling to select only the most likely tokens.} 

Classifications that vary across iterations may be “edge cases'' and have  a lower probability of correct classification.\footnote{Accessing token log probabilities directly, once available, will be an effective way to a similar analysis.}  This approach rests on the core assumption that the full distribution of token probabilities captures latent information about the annotation’s classification. If, for example, the top tokens are similar in probability, then choosing one of these tokens may misrepresent the model’s annotation decision. Instead, measuring the variability across iterations allows us to find these “edge cases.” We call this measure of uncertainty in the annotation label a “consistency score.'' We define an indicator function $C(i)$ that is equal to 1 when label $i$ for a given task is equal to the LLM's modal classification, $m$, for task :
\begin{equation*}
    C(i)=
    \begin{cases}
      1 & \text{if}\ i=  m \\
      0 & \text{otherwise}
    \end{cases}
\end{equation*}
Given a vector of classifications, $\mathbf{a}$, with length $l$ for a given classification task, \textit{consistency} is measured as the proportion of classifications that match the modal label:
\begin{equation*}
    Consistency = \frac{1}{l}\sum_{j=1}^l C(a_j)
\end{equation*}

For these ablation experiments, we classify every text sample three times at a temperature of 0.7 and measure each text sample’s consistency score. Because there are only three iterations, each text sample can only have two values for consistency score: 0.67 and 1.0. Across all analyzed tasks, classifications with a consistency of 1.0 show significantly higher accuracy (19.4\% increase), true positive rate (16.4\% increase), and true negative rate (21.4\% increase) compared to classifications with a consistency less than 1.0. Roughly 85\% of classifications had a consistency of 1.0.

\begin{table*}[b] 
\centering
\begin{tabular}{p{4 cm}p{3.5 cm}p{3.5cm}p{2 cm}}
\toprule
\textbf{Data set and task}             & \textbf{BERT F1 score (training obs w/o noise)} & \textbf{BERT F1 score (training obs w/ noise)} & \textbf{Difference} \\ 
\midrule
Hopkins (AJPS): Political    & 0.340                 & 0.344               & -0.004        \\
\midrule
Hopkins (AJPS): religion     & 0.609                 & 0.609               & 0.000         \\
\midrule
Hopkins (AJPS): gender       & 0.716                 & 0.684               & 0.032         \\ 
\midrule
Hopkins (AJPS): race         & 0.635                 & 0.640               & -0.005        \\ 
\midrule
Muller (JOP): future         & 0.851                 & 0.851               & 0.000         \\
\midrule
Muller (JOP): past           & 0.791                 & 0.755               & 0.036         \\ 
\midrule
Muller (JOP): present        & 0.606                 & 0.601               & 0.005         \\ 
\midrule
Card (PNAS): cat\_imm        & 0.832                 & 0.815               & 0.017         \\ 
\midrule
Card (PNAS): cat\_anti       & 0.565                 & 0.573               & -0.008        \\ 
\midrule
Card (PNAS): cat\_neutral    & 0.385                 & 0.428               & -0.043        \\ 
\midrule
Card (PNAS): cat\_pro        & 0.448                 & 0.436               & 0.012         \\
\midrule
Peng (PNAS)                  & 0.431                 & 0.444               & -0.013        \\ 
\midrule
Saha (PNAS): CV              & 0.031                 & 0.059               & -0.028        \\ 
\midrule
Saha (PNAS): HD              & 0.210                 & 0.276               & -0.066        \\ 
\bottomrule
\end{tabular}
\caption{Comparing BERT F1 score for models fine-tuned with and without noise}
\label{tab:noise}
\end{table*}

Table \ref{tab:noise} shows supervised model performance for BERT models fine-tuned on 1,250 training observations labeled by GPT-4 (i.e., labels with noise) compared to BERT models fine-tuned on training observations with a consistency score of 1.0 (i.e., labels without noise), which reduced our training set to slightly more than 1000 samples per task. Put otherwise, the second series of models involved dropping about 250 text samples per task so that the training set only retained annotations where GPT-4 consistently labeled the same category across all iterations. 

Our findings indicate that there are minimal differences between models trained on labels with noise and labels without noise. Models trained without noise display, on average, 0.004 lower F1 score than models trained with noise. These results suggest that the supervised models explored here are fairly robust to noise in the labels.

\subsection{Comparing GPT-4 few-shot performance over time}

Our final set of ablation experiments involved replicating the GPT-4 few-shot model at different points in time. An unsettling scenario involves the potential drift in capabilities as generative LLMs undergo opaque changes and updates. Some research, such as \citet{lingjiao}, claim that GPT-4 performance is declining over time. To account for the potential of changing model weights in GPT-4, we re-analyzed each task six months after our initial analyses and compared results across time. 

\begin{figure*}
    \centering
    \includegraphics[width = .9\textwidth]{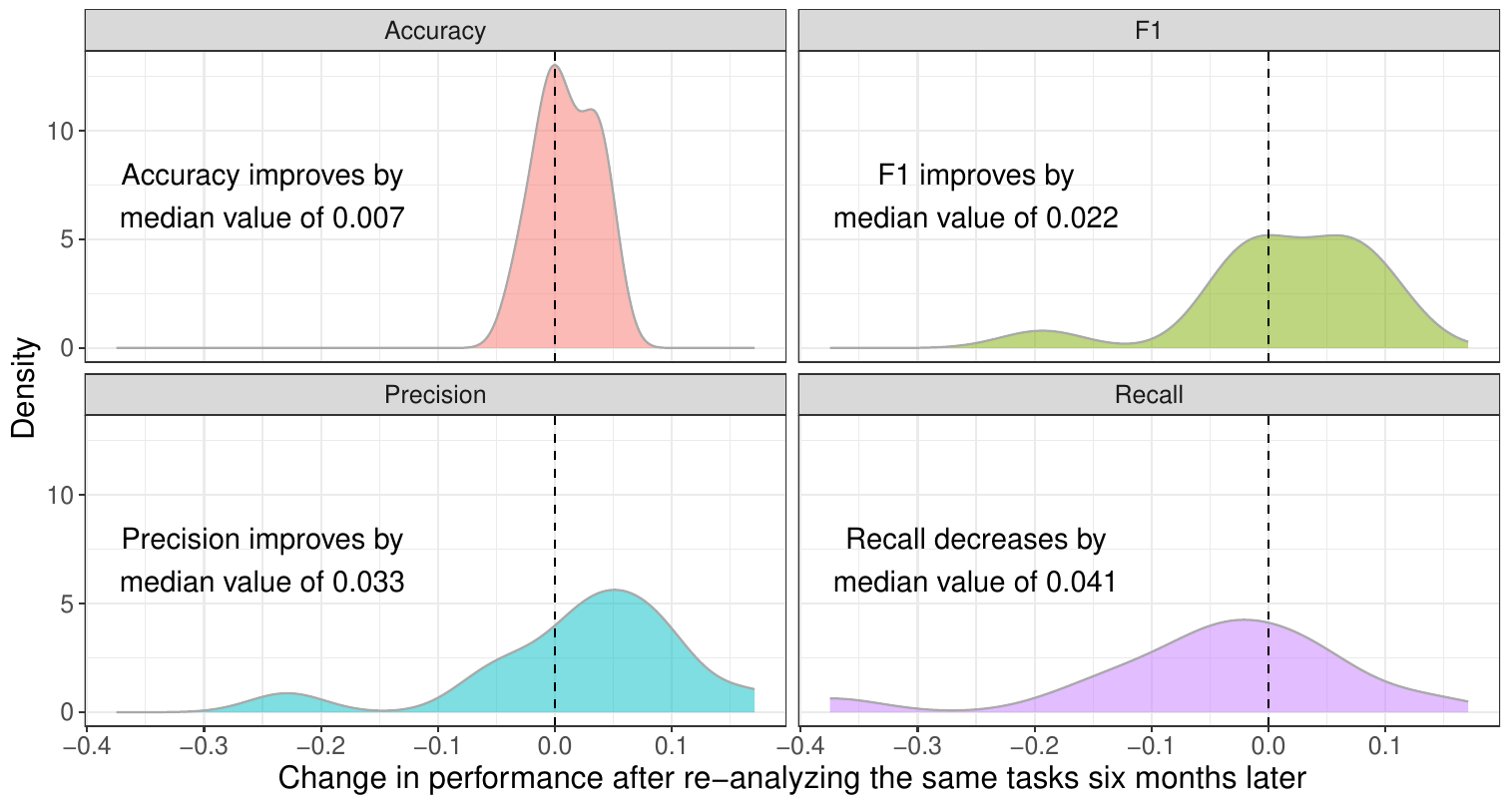}
    \caption{Examining GPT-4 performance over time}
    \label{fig:overtime}
\end{figure*}

Figure \ref{fig:overtime} shows evaluation comparisons of few-shot tasks in both April 2023 and November 2023. Our results do not suggest significant changes in GPT-4 performance over time. If anything, Figure \ref{fig:overtime} reveals a small \textit{increase} in performance since my initial experiments. Across the 14 tasks, accuracy improved by 0.007 and F1 increased by 0.022 when the same annotation procedures were carried out in November 2023.

\section{Appendix: Miscellaneous additional information}

Additional sources:
\begin{itemize}
    \item Robot image (used in Figure 1): \url{https://commons.wikimedia.org/wiki/File:Grey_cartoon_robot.png}
    \item Human silhouette image (used in Figure 1): \url{https://commons.wikimedia.org/wiki/File:SVG_Human_Silhouette.svg}
\end{itemize}

\end{document}